\documentclass[10pt,twocolumn,letterpaper]{article}

 \usepackage{cvpr}              % To produce the CAMERA-READY version
\definecolor{cvprblue}{rgb}{0.21,0.49,0.74}
\usepackage[pagebackref,breaklinks,colorlinks,allcolors=cvprblue]{hyperref}
\usepackage{graphics}
\usepackage{multirow}
\usepackage{makecell}
\usepackage{ragged2e}
\usepackage{float}
\usepackage{caption}

\title{3SGen: Unified Subject, Style, and Structure-Driven Image Generation with Adaptive Task-specific Memory}

\author{Xinyang Song\textsuperscript{\rm 1, 2}, Libin Wang\textsuperscript{\rm 3}, Weining Wang\textsuperscript{\rm 2}, Zhiwei Li\textsuperscript{\rm 1, 2}, Jianxin Sun\textsuperscript{\rm 3}, Dandan Zheng\textsuperscript{\rm 3}, \\
Jingdong Chen\textsuperscript{\rm 3}, Qi Li\textsuperscript{\rm 1, 2}, Zhenan Sun\textsuperscript{\rm 1, 2}\\
\textsuperscript{\rm 1}School of Artificial Intelligence, UCAS \quad
\textsuperscript{\rm 2}CASIA \quad \textsuperscript{\rm 3}AntGroup\\
}

\begin{document}

\twocolumn[{
\maketitle
\begin{center}
    \vspace{-18pt}
    \captionsetup{type=figure}
    \includegraphics[width=0.96\textwidth]{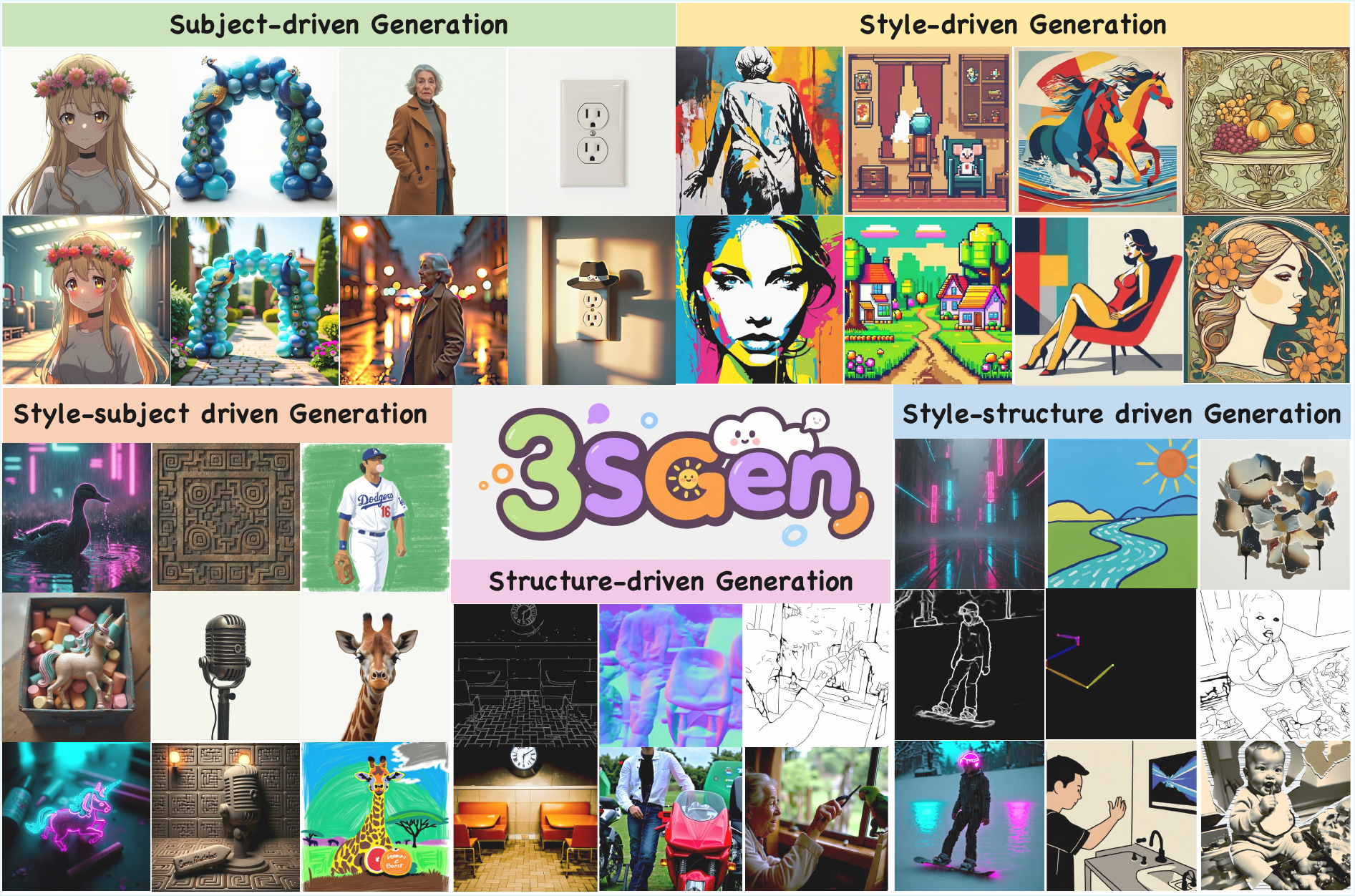}
    \vspace{-5pt}
    \captionof{figure}{Versatile samples of the proposed \textbf{3SGen}. Given one or multiple reference images along with a text prompt, 3SGen generates consistent high-fidelity results.}
%    \vspace{-5pt}
    \label{fig:1}
\end{center}
}]

\begin{abstract}

Recent image generation approaches often address subject, style, and structure-driven conditioning in isolation, leading to feature entanglement and limited task transferability.
In this paper, we introduce \textbf{3SGen, a task-aware unified framework that performs all three conditioning modes within a single model.} 3SGen employs an MLLM equipped with learnable semantic queries to align text–image semantics, complemented by a VAE branch that preserves fine-grained visual details. At its core, an \textbf{Adaptive Task-specific Memory (ATM) module} dynamically disentangles, stores, and retrieves condition-specific priors, such as identity for subjects, textures for styles, and spatial layouts for structures, via a lightweight gating mechanism along with several scalable memory items. This design mitigates inter-task interference and naturally scales to compositional inputs. In addition, we propose \textbf{3SGen-Bench}, a unified image-driven generation benchmark with standardized metrics for evaluating cross-task fidelity and controllability. Extensive experiments on our proposed 3SGen-Bench and other public benchmarks demonstrate our superior performance across diverse image-driven generation tasks.

\end{abstract}    
\section{Introduction}
\label{sec:intro}

Image generation has evolved rapidly from simple text-to-image synthesis to controllable generation guided by visual conditions. 
Recent studies can be broadly categorized into three major paradigms based on the type of conditions, namely subject-driven personalization~\cite{xieomnicontrol, ye2023ip, ruiz2023dreambooth, labs2025flux, galimage, wu2025less}, style-driven transfer~\cite{qi2024deadiff, xing2024csgo, han2024stylebooth, li2024styletokenizer, wang2024instantstyle, frenkel2024implicit}, and structure-driven~\cite{mou2024t2i, zhang2023adding, huang2023composer} generation.
While each line of research has achieved remarkable success, these tasks are typically developed in isolation, overlooking their intrinsic complementarity. 
In practice, emphasizing one aspect of generation often imposes constraints on the others. Preserving subject identity necessitates suppressing stylistic variations, refining style transfer demands degrading instance-specific content, and enforcing structural control prioritizes spatial composition over appearance or identity.
Analogous to the human visual system, where distinct regions and neurons process color, texture, and shape yet operate jointly to yield coherent perception, image generation tasks involving subject, style, and structure conditioning demand both shared and independent representations.
By leveraging these interdependencies, tasks can be better disentangled, improving feature modularity and cross-task generalization.

Recent efforts in unified multimodal models (UMMs) ~\cite{chen2025blip3, gao2025seedream, labs2025flux, lin2025uniworld, wu2025omnigen2, wu2025qwen} aim to integrate textual and visual conditions within a single architecture. 
While these models offer scalability across diverse tasks, they primarily emphasize architectural unification without explicitly addressing the need for task-aware integration.
Consequently, existing methods tend to suffer from entangled feature representations and weak cross-task performance, as they fail to properly disentangle and adapt the representations according to the unique characteristics of each conditioning type.
Furthermore, some other frameworks~\cite{xing2024csgo, wu2025uso} employ separate visual encoders to handle heterogeneous conditions, but this architectural split introduces semantic gaps and hinders cross-task generalization.

Motivated by the above observations, we propose 3SGen, a unified framework that integrates subject, style, and structure conditioning, enabling flexible and task-specific image generation across diverse conditions.
Unlike prior unified models that primarily merge architectures, 3SGen achieves task-aware unification, enabling different conditioning types to interact seamlessly while minimizing task-related conflicts and benefiting from shared representations.
Our design is built upon two key insights: (1) a unified generator must capture both modality alignment across text and image inputs and task differentiation across conditioning types; and (2) task-specific knowledge should be dynamically retrievable rather than statically encoded.

To this end, we employ a multimodal large language model (MLLM) as a semantic bridge that jointly encodes textual and visual information. A set of learnable semantic queries functions as feature resamplers to align the modalities, while an auxiliary visual encoder complements high-frequency visual details. Beyond alignment, we introduce an Adaptive Task-specific Memory (ATM) module that learns to retain, retrieve, and integrate task-relevant priors. Each memory item encodes distinctive semantics, including identity for subjects, texture for styles, and spatial priors for structures. An adaptive gating mechanism dynamically selects and fuses relevant memories according to the conditioning type. This conditional adaptation mitigates inter-task interference, enhances controllability, and provides straightforward scalability to new condition types. 

%The framework of 3SGen is shown in Fig.~\ref{fig:1}.

% Although such unified modeling is essential, the community still lacks a standardized benchmark to evaluate how well a single model handles diverse conditioning modes. Existing datasets assess each task separately, using inconsistent metrics and protocols. To fill this gap, we establish 3SGen-Bench, the first benchmark dedicated to unified image-driven generation across subject, style, and structure based conditions.
% 3SGen-Bench provides a standardized data protocol and comprehensive evaluation suite to jointly assess fidelity, controllability, and cross-task generalization.
% Specifically, it includes three complementary subsets focusing on (1) subject identity preservation, (2) style similarity and consistency, and (3) structural alignment and spatial coherence, together with unified quantitative metrics.
% This benchmark enables fair comparison among heterogeneous methods and serves as a foundation for analyzing the interplay between different conditioning paradigms within a single unified framework.

Through this unified yet adaptive paradigm, 3SGen bridges previously disjoint generation tasks and establishes a practical pathway toward general-purpose image generation.
To facilitate comprehensive evaluation and foster progress in universal image-driven generation, we propose 3SGen-Bench, a unified multi-task benchmark specifically designed to assess model performance across diverse image-guided generation scenarios.
In addition to diverse subject and style references, 3SGen-Bench introduces structure references, accompanied by comprehensive textual prompts. 
To address the inadequacy of existing metrics, 3SGen-Bench leverages MLLM for multi-dimensional evaluation of multimodal alignment across multiple tasks.
Extensive experiments on both 3SGen-Bench and public benchmarks~\cite{pengdreambench++, labs2025flux} validate the effectiveness of our approach, showcasing significant improvements in controllability, fidelity, and cross-task generalization.
The main contributions of this paper can be summarized as follows.

\begin{itemize}

\item[$\bullet$] 
%To our knowledge, 3SGen represents the first unified image generation framework capable of performing subject, style, and structure-driven synthesis within a single model under multimodal inputs.
We propose a \textbf{task-aware unified framework 3SGen} that explicitly disentangles and adapts to \textbf{s}ubject, \textbf{s}tyle, and \textbf{s}tructure-driven generation within a single model.
\item[$\bullet$] 
We introduce \textbf{Adaptive Task-specific Memory (ATM)} module, a task-aware memory module that dynamically stores and retrieves condition-specific prior, along with an adaptive gating mechanism that adaptively disentangles representations for each task. This lightweight module is designed to be plug-and-play with flexible scalability.
%thereby refining the distribution of condition queries.
\item[$\bullet$] 
We establish \textbf{3SGenBench}, a comprehensive benchmark for cross-task image-driven generation. Extensive experiments across diverse scenarios demonstrate the superior performance and generalizability of our proposed model.

\end{itemize}

\section{Related Works}
\label{sec:related works}

\subsection{Image-Driven Generation}
%Early work like DALLE-2~\cite{ramesh2022hierarchical} learns semantic correspondences between images and textual descriptions during training via CLIP-based image-text alignment.
%The recent powerful text-to-image generation base models, like Stable Diffusion~\cite{rombach2022high, esser2024scaling} and FLUX~\cite{Flux}, along with controllable plugins built upon them, have significantly improved the convenience and effectiveness of performing this task.
%Through careful regularization and the introduction of unique textual identifiers, Textual Inversion~\cite{galimage} and DreamBooth~\cite{ruiz2023dreambooth} teach pre-trained text-to-image diffusion models new visual concepts using a set of reference images.
%Recently, transformer-based models become increasingly prevalent in image-guided generation, owing to their powerful in-context learning abilities.
%, which significantly advance research in image-driven synthesis. 
%Approaches such as ICLoRA~\cite{huang2024context}, OmniControl~\cite{xieomnicontrol}, UNO~\cite{wu2025less}, and FLUX.1-Kontext~\cite{labs2025flux} utilize shared attention mechanisms to maintain the in-context consistency between the generated and reference images, enabling the text-to-image models adapted for subject-driven tasks.
%While existing approaches predominantly focus on single-type image guidance, 3SGen unifies multiple tasks and enables general-purpose image-driven generation.
%This unified approach overcomes the limitations of previous methods, providing more flexibility and controllability over the generated images.

Image-driven generation extends text-to-image synthesis by incorporating image-specific conditions, providing finer control over the generation process. 
Early methods such as ControlNet~\cite{zhang2023adding} and T2I-Adapter~\cite{mou2024t2i} introduce structural control by introducing a trainable copy of the diffusion network, enabling manipulation of pre-trained models.
Other approaches, including Textual Inversion~\cite{galimage} and DreamBooth~\cite{ruiz2023dreambooth}, teach pre-trained text-to-image diffusion models new visual concepts using sets of reference images.
Additionally, IP-Adapter~\cite{ye2023ip} enables flexible image personalization with a decoupled cross-attention adapter, obviating the need for expensive fine-tuning.
Recently, transformer-based models such as ICLoRA~\cite{huang2024context}, OmniControl~\cite{xieomnicontrol}, UNO~\cite{wu2025less}, and FLUX.1-Kontext~\cite{labs2025flux} leverage shared attention mechanisms to maintain consistency between generated and reference images, demonstrating strong performance on subject-driven generation tasks.
However, these methods primarily focus on a single conditioning type. To address these limitations, 3SGen unifies multiple conditioning types and enables general-purpose image-driven generation across subject, style, and structure. 

\subsection{Unified Multimodal Models}

%Recent advances in vision-language models~\cite{Gemini, Gpt4o, chen2025blip3} have spurred significant progress in unified multi-modal models (UMMs), fostering the development models capable of performing a wide range of generative tasks. 
%UMMs integrate both generation and understanding capabilities into a single framework, allowing for general-purpose multi-modal generation, including image synthesis, image understanding, and other creative tasks.
%unified multi-modal models
%Early works~\cite{wang2024emu3, team2024chameleon, xie2024show, wu2025janus} integrate visual encoders with large language models and perform autoregressive decoding to enable visual synthesis.
%Recent approaches~\cite{xiao2025omnigen, zhang2025nexus,  huang2025illume+, gao2025seedream, xie2025show, deng2025emerging} pursue multi-task frameworks for broadly general visual generation. 
%More efforts~\cite{pan2025transfer, lin2025uniworld, wu2025omnigen2, wu2025qwen} adopt hybrid architectures to combine vision-language models with diffusion backbones, thus integrating semantic reasoning with pixel synthesis.

Recent advances in UMMs~\cite{Gemini, Gpt4o, chen2025blip3, wang2024emu3, team2024chameleon, xie2024show, wu2025janus, xiao2025omnigen, zhang2025nexus,  huang2025illume+, gao2025seedream, xie2025show, deng2025emerging, pan2025transfer, lin2025uniworld, wu2025omnigen2, wu2025qwen} have substantially enhanced vision-language integration, enabling models to perform a wide range of generative tasks.
Early works, such as Chameleon~\cite{team2024chameleon} and Show-O~\cite{xie2024show} integrate visual encoders with large language models (LLMs) and use autoregressive decoding for visual synthesis. 
Recent approaches, including OmniGen~\cite{xiao2025omnigen}, Nexus-Gen~\cite{zhang2025nexus}, and BAGEL~\cite{deng2025emerging}, pursue multi-task frameworks for general visual generation. 
More recent efforts, such as UniWorld~\cite{lin2025uniworld}, OmniGen2~\cite{wu2025omnigen2} and Qwen-Image~\cite{wu2025qwen} combine vision-language models (VLMs) with diffusion backbones to integrate semantic understanding with pixel synthesis. While these models achieve improved performance across various conditions, they still inadequately decouple conditioning factors such as subject identity, style, and structure, leading to conflicts when combining these conditions.
In contrast, 3SGen introduces a task-aware unification framework featuring an Adaptive Task-specific Memory (ATM) module. This module dynamically retrieves and updates task-specific priors for each conditioning type, enabling independent modulation and decoupling of conditions, thereby enhances flexibility, controllability, and performance in complex multi-condition scenarios.
%However, these approaches simply unify multi-task training without effectively decoupling or modulating different conditioning types, leading to entangled feature representations. In contrast, 3SGen addresses this challenge with task-aware unification, providing more flexibility and controllability across multiple conditions.
%offering better control and separation across multiple conditions.

\subsection{Multi-condition Modular Approaches}
%\subsection{Modular and Adaptive Approaches}
%Memory-based and modular approaches have emerged as a promising strategy to handle diverse conditional inputs in generative models.
%UniControl~\cite{qin2023unicontrol} unifies multiple types of control signals by designing versatile control modules.
%While Unic-Adapter~\cite{duan2025unic} presents a plug-and-play adapter framework for multi-conditional generation, it still relies on specific instructions as condition category signals.
%Recent effort like USO~\cite{wu2025uso} leverages separate encoders for style and subject control, yet does not incorporate structural information. This inconsistent encoding strategy often leads to semantic misalignment and decreases overall consistency.
Recent works like UniControl~\cite{qin2023unicontrol}, Unic-Adapter~\cite{duan2025unic}, and USO~\cite{wu2025uso} have introduced modular frameworks to handle multi-conditional generation tasks. These models employ modular components, such as adapters or separate visual encoders, to handle different types of conditioning. While these approaches offer flexibility across multiple tasks, they do not effectively address the challenge of decoupling conditioning factors or modulating task-specific features.
In this work, 3SGen leverages its ATM module to dynamically retrieve and integrate task-specific priors, effectively disentangling task features while maintaining shared representations, offering a more scalable solution.
%for multi-condition generation.
\section{Method}

\begin{figure*}
  \centering
  \vspace{-12pt}
  \includegraphics[width=0.95\linewidth]{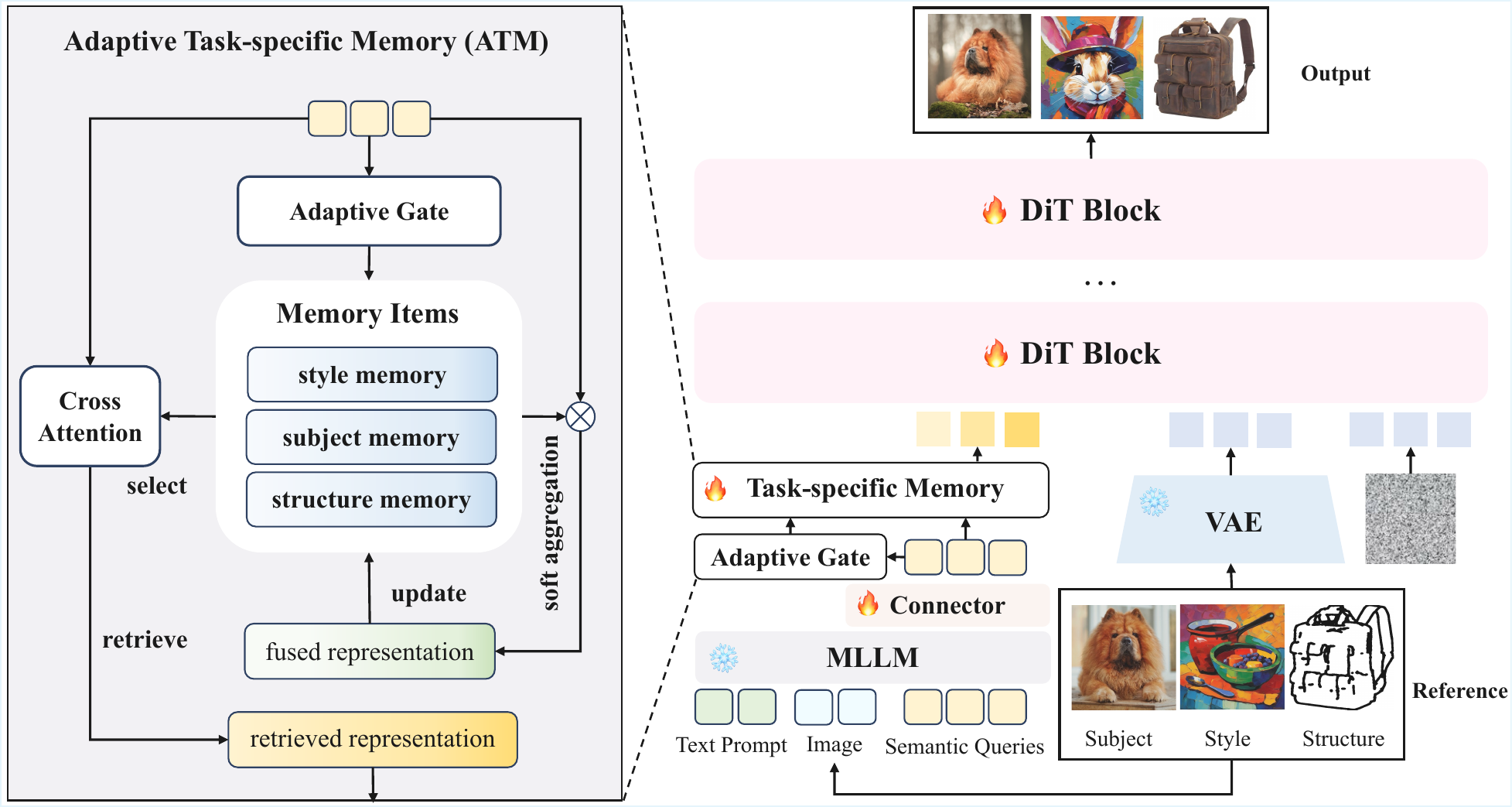}
    \vspace{-5pt}
    \caption{The overall architecture of our proposed 3SGen. Semantic queries serve as resamplers that pass through the MLLM to obtain aligned multimodal features, which then interact with adaptive task-specific memory priors to derive task-aware semantic representations. VAE tokens from reference images are incorporated as joint inputs to supplement fine-grained visual details.}
    \vspace{-10pt}
    \label{fig:2}
\end{figure*}

%In this work, we present 3SGen, a unified framework for subject, style, and structure-driven image generation that introduces aligned multi-modal conditional encoding while preserving the high-quality visual synthesis capabilities of diffusion models. 
%In Sec.~\ref{3.1}, we present the unified design of the 3SGen framework (Fig.~\ref{fig:2-1}). 
%In Sec.~\ref{3.2}, we design a task-specific memory module for multi-category conditions along with an adaptive gate (Fig.~\ref{fig:2-2}). 
%In Sec.~\ref{3.3}, we design a hierarchical training strategy and introduce the details of training recipe.

As shown in Fig.~\ref{fig:2}, we present the overall design of 3SGen, a unified framework for subject, style, and structure-driven image generation. 
In this section, we first describe how 3SGen achieves unified multimodal conditioning through semantic alignment facilitated by an MLLM (Sec.~\ref{3.1}). Then, we describe the Adaptive Task-specific Memory (ATM) module, which enhances task-aware modulation (Sec.~\ref{3.2}). Finally, we outline the hierarchical training progress employed to optimize the framework (Sec.~\ref{3.3}).

\subsection{Unified Image-driven Generation}
\label{3.1}

Building upon the motivation outlined in Sec.~\ref{sec:intro}, 
our goal is to design a unified conditional generator that seamlessly integrates subject, style, and structure-based controls within a single model, while maintaining robust text-image alignment and high-fidelity visual synthesis. 
Existing methods~\cite{xiao2025omnigen, xieomnicontrol, wu2025less, wu2025uso, duan2025unic, labs2025flux} separately encode visual and textual tokens, enabling cross-modal interactions solely through the DiT attention layers.  However, as demonstrated in previous works~\cite{yao2025reconstruction, yurepresentation}, the image denoising-based modeling paradigm of DiT reveals limitations in its capacity for multimodal semantic understanding.
%DiT's image denoising-based modeling paradigm exhibits limited capability in multi-modal semantic understanding. 
%is inherently suited for visual generation but 
This independent encoding strategy often leads to conflicts between visual and textual conditions, thereby reducing the coherence and visual quality of the generated images. 
%In contrast, approaches like OmniGen2~\cite{wu2025omnigen2} and Qwen-Image~\cite{wu2025qwen} utilize the latent hidden states from the final layer of Multi-modal Large Language Models (MLLMs) as condition. Nevertheless, since the overall parameters of the MLLM remain frozen during training, the representations derived from these hidden states often fail to capture rich semantic content in the absence of trainable parameters.

To address these issues, 3SGen adopts a multimodal large language model (MLLM) as a semantic bridge, 
jointly encoding textual and visual conditions into a shared latent space. 
Previous approaches~\cite{wu2025omnigen2, wu2025qwen} utilize the final hidden states of MLLM as the conditional representation, which often fail to capture the full semantic richness due to the lack of trainable parameters. 
To overcome this limitation, we introduce a set of learnable semantic queries serving as dynamic resamplers of cross-modal representations. 
%that propagate through every layer of the MLLM,  serving 
Specifically, we input these learnable semantic queries alongside text and image tokens into the MLLM, and map the resulting output through a connector to the DiT input space.
During training, both the semantic queries and the connector are jointly optimized, while the MLLM remains frozen. 
By propagating through all layers of the MLLM, these learnable queries can capture richer cross-modal semantics compared to relying solely on the final hidden states.
%rich semantic content in the absence of trainable parameters.
%However, directly using the frozen hidden states of MLLMs (as done in OmniGen2~\cite{wu2025omnigen2} or Qwen-Image~\cite{wu2025qwen}) often fails to capture fine-grained, task-relevant semantics.

Complementary to this semantic alignment, we incorporate a variational autoencoder (VAE) %visual tokenizer 
to compress each input image into compact latent embeddings. 
These embeddings complement the high-level semantics from the MLLM with 
fine-grained textures and structural cues that are often underrepresented 
in purely semantic spaces. 
This joint design achieves both \textit{modality unification} across text and image inputs 
and \textit{semantic preservation} of detailed visual information, 
thus laying a strong foundation for high-fidelity image-driven generation.
%the subsequent Adaptive Task-specific Memory module that performs task-aware disentanglement and compositional fusion (Sec.~\ref{3.2}).

%Image-driven generation builds upon text-to-image synthesis by incorporating image-based conditions. The key challenge lies in effective semantic extraction and feature fusion from both textual and visual modalities, ensuring that accurate and coherent guidance is provided for the subsequent denoising process. 

%To obtain aligned multi-modal encodings, we adopt an MLLM with learnable queries approach similar to MetaQuery~\cite{pan2025transfer}, leveraging the MLLM as a feature resampler while introducing learnable parameters for efficient representation learning.  Specifically, we input fixed-length learnable semantic queries alongside text and image tokens to the MLLM, and map the corresponding output through a trainable connector to the DiT input space.
%During training, both the semantic queries and the connector are jointly optimized, while the MLLM is kept frozen.  By propagating through every layer of the MLLM, these learnable queries can capture richer cross-modal semantics compared to solely relying on the final hidden states.
%The computation of the multi-modal condition $C$ proceeds as follows:
%In addition, recognizing that MLLMs primarily focus on visual-semantic information, we employ a variational autoencoder (VAE) as a visual tokenizer to compress the images into compact latent representations, serving as a complement to visual conditions for preserving fine-grained image details.

\subsection{Adaptive Task-specific Memory Module}
\label{3.2}

In practical applications, image conditions often exhibit significant category-specific variations. 
Treating all input conditions uniformly can lead to homogeneous condition distributions, thereby diminishing the model’s specialization and effectiveness.
Intuitively, subject conditions emphasize the extraction of content-relevant semantics, style conditions focus on capturing global color schemes and fine-grained textures, while structure conditions primarily address the overall spatial layout.
%UnicAdapter~\cite{duan2025unic} handles multi-category conditions through network structure duplication. 
%However, it relies on additional textual instructions and lacks support for multi-condition compositional generation. 
%USO~\cite{wu2025uso} employs VAE and SigLIP encoders to separately extract subject and style features. 
Existing approaches~\cite{duan2025unic, wu2025uso} handle multi-category conditions through network structure duplication or employing separate vision encoders, which neither enable feature disentanglement nor scale to a broader range of condition types.
To address these limitations, we propose an Adaptive Task-specific Memory Module for unified multi-task condition modulation.

\noindent \textbf{Task-specific Memory.}
Building on the multimodal semantic inputs introduced in Sec.~\ref{3.1}, 
%where we introduced the multimodal semantic inputs, 
we now focus on deriving task-specific semantic conditions tailored to distinct generation objectives.
% The conditional queries introduced in Sec.~\ref{3.1} serve as unified multi-modal semantic inputs, while our objective is to derive task-specific semantic conditions tailored for distinct generation goals. 
Inspired by~\cite{huang2021memory, sun2024anyface++}, we propose the task-aware memory module to capture the distribution of different conditional queries, enabling dynamic modulation during inference, as illustrated in Fig.~\ref{fig:2}.
Specifically, the memory module comprises $n$ task-specific memory items $\mathcal{M} \in \mathbb{R}^{m \times c}$, where $m$ denotes the memory capacity, and $c$ is the channel dimension.
For a given semantic query $Q \in \mathbb{R}^{l \times c}$, we compute cross-attention between $Q$ and the corresponding $\mathcal{M}_i$, $(i = 1, ..., n)$, yielding a retrieved representation $R \in \mathbb{R}^{l \times c}$ and an attention weight matrix $W \in \mathbb{R}^{l \times m}$ via self-supervised retrieval:
\begin{equation}
    R = \mathrm{Attn}(Q, \mathcal{M}_i); \; \; \; W = \text{Softmax}\left( \frac{Q \mathcal{M}_i^\top}{\sqrt{c}} \right).
\end{equation}
Here, the retrieved representation $R$ serves as the distribution-modulated conditional queries which are then fed into DiT blocks, while the attention weight matrix $W$ quantifies the similarity scores between each token in memory items and the semantic queries.
%During training, the retrieved representations are fused with input queries through soft aggregation, weighted by their similarity scores $W$:
During training, the input queries, weighted by the similarity scores $W$, are fused with the memory item through soft aggregation, :
\begin{equation}
    \mathcal{\hat{M}}_i = \alpha\cdot(\frac{W}{\sum_{j} W_{j,:}})^\top Q + (1-\alpha)\cdot \mathcal{M}_i ,
\end{equation}
where $\mathcal{\hat{M}}_i$ denotes the fused representation to update $\mathcal{M}_i$, and $\alpha$ is the exponential moving average coefficient.
This process facilitates continuous updating of each task-specific memory item, allowing the model to dynamically store the information of each semantic query.
During inference, the memory items are fixed without updates. Through dynamic retrieval and updating, the model learns distinct task-specific conditional distribution priors and performs biased modulation on the semantic queries.

Moreover, 3SGen naturally supports multi-condition compositional generation, such as \textit{style-subject} and \textit{style-structure} driven generation. 
After modulation through multiple memory items, retrieved queries are concatenated along the sequence dimension to form compositional semantic queries, with VAE embeddings concatenated accordingly. The multi-condition results are shown in Fig.~\ref{fig:9}.

\noindent \textbf{Adaptive Gate.}
To enable condition category awareness, we introduce an adaptive gate that discriminates among condition types, allowing the model to automatically select the appropriate memory module for retrieval and updates. 
Specifically, the adaptive gate is implemented as a multi-class MLP, which takes the semantic queries $Q$ as input, predicts the current task type, and provides corresponding directive signals to the memory module.
By integrating the adaptive gate with the memory module, we establish a scalable and unified method for conditioning adaptation. This framework can be easily extended to accommodate additional condition categories by simply adjusting the number of memory items and the classification capacity of the gate.

\begin{figure}[t]
    \centering
    \vspace{-12pt}
    \includegraphics[width=1.0\linewidth]{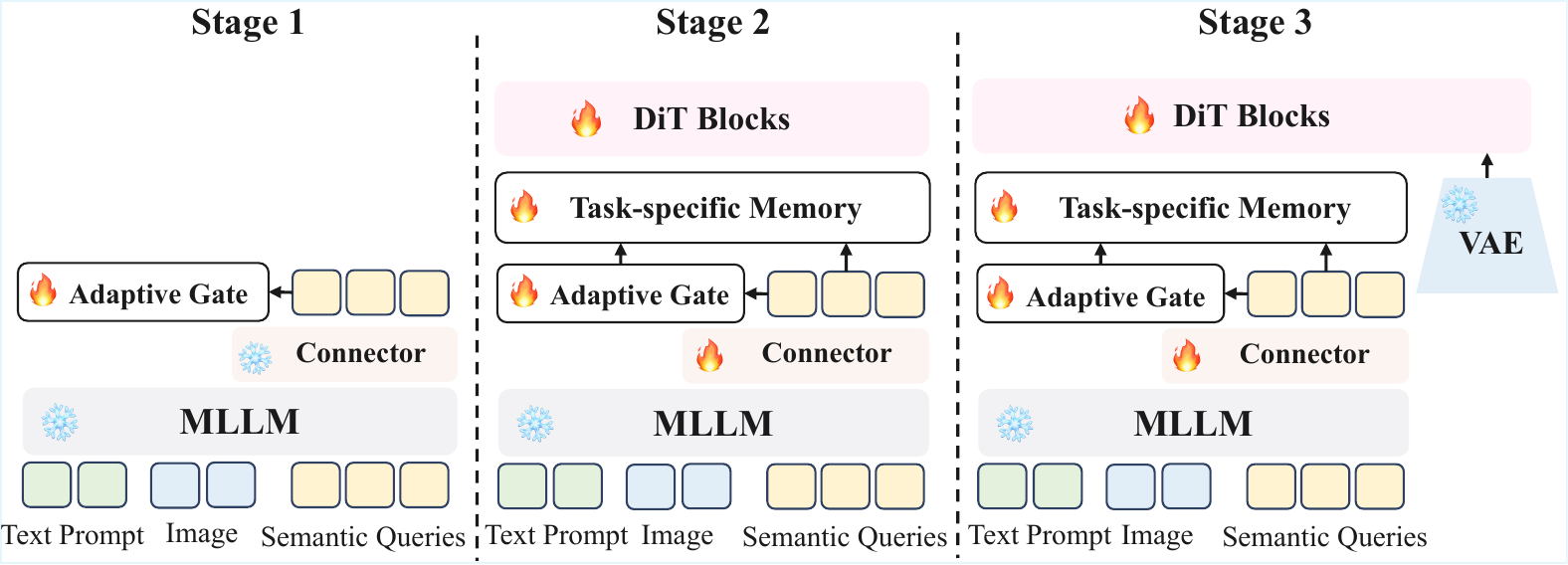}
    \vspace{-12pt}
    \caption{\textbf{Visualization of the hierarchical training strategy.} Noise input is omitted for simplification.}
    \vspace{-10pt}
    \label{fig:3}
\end{figure}

\subsection{Training Schedule}
\label{3.3}
Building on the framework introduced in the previous sections, we propose a multi-stage hierarchical training strategy to ensure semantically consistent and high-fidelity generation results, as illustrated in Fig.~\ref{fig:3}. 
The first stage involves pretraining the adaptive gate, during which only the parameters of the gate are updated to initialize precise condition type discrimination.
We found that directly incorporating VAE features with semantic queries may lead to degraded text-image consistency.
%To address this, our generation training proceeds in phases: the second stage training is conditioned only on semantic queries, followed by a joint training phase in which both semantic queries and VAE features are included. This hierarchical approach enables rapid convergence on textual semantics, particularly crucial for structure-driven scenarios.
To address this, our generation training proceeds in phases: the second stage involves training conditioned solely on semantic queries, followed by a joint training phase that includes both semantic queries and VAE features. This hierarchical approach accelerates convergence on textual semantics, which is particularly crucial for structure-driven scenarios.

Our mixed training data is sourced from various datasets: subject-driven data from Subject200K~\cite{xieomnicontrol} and UNO~\cite{wu2025less}, style-driven data including StyleBooth~\cite{han2024stylebooth} images and our custom collection, and structure-driven data curated from UniWorld~\cite{lin2025uniworld}. For multi-condition compositional scenarios, we construct triplet data through instruction-guided style transfer. Diverse style descriptions are generated using GPT-5~\cite{Gpt5}, and edited images are created via Gemini-Flash~\cite{comanici2025gemini}, producing 50K image pairs each for subject-style and structure-style driven generation.

\begin{figure}[t]
    \centering
    \vspace{-12pt}
    \includegraphics[width=0.98\linewidth]{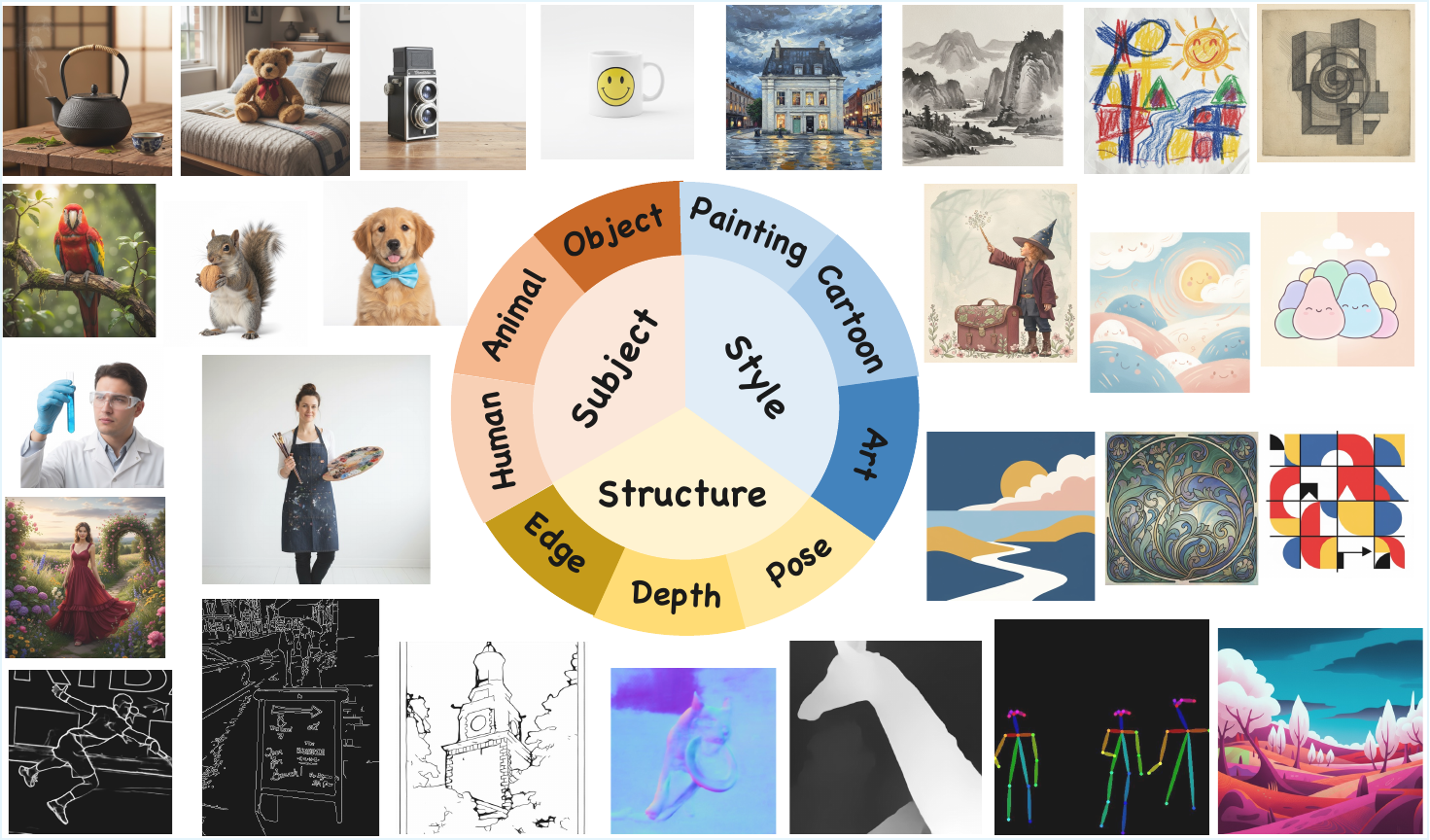}
    \vspace{-8pt}
    \caption{Showcases of samples from our proposed 3SGen-Bench.}
    \vspace{-12pt}
    \label{fig:4}
\end{figure}

\section{Experiments}

\begin{table*}
    \centering
    \vspace{-12pt}
    \caption{\textbf{Quantitative evaluation of multi-task image-driven generation.} The best results are highlighted in bold, while the second-best results are marked with an underline.}
    \vspace{-5pt}
    \label{table:1}
{\footnotesize %\small
\begin{tabular}{lcccccccccc}
\toprule
\multirow{2}{*}{Method}     & \multicolumn{4}{c}{Subject-driven Generation}    & \multicolumn{3}{c}{Style-driven Generation} & \multicolumn{3}{c}{Structure-driven Generation}
\\   \cmidrule(r){2-5}   \cmidrule(r){6-8}  \cmidrule(r){9-11}
& CLIP-I$\uparrow$ & DINO$\uparrow$ & CLIP-T$\uparrow$ & FID$\downarrow$ & CSD$\uparrow$ & CLIP-T$\uparrow$ & FID$\downarrow$ & Struc-Sim$\uparrow$ & CLIP-T$\uparrow$ & FID$\downarrow$ \\ 
\midrule
OmniGen2~\cite{wu2025omnigen2}       & 0.620  & 0.735  & 0.288 & 138.04 & 0.471 & 0.279 & 152.89 & - & - & -  \\ 
Flux Kontext dev~\cite{labs2025flux} & \underline{0.642} & \underline{0.757}  & 0.285  & 136.58 & 0.506 & 0.280 & \textbf{132.92} & - & - & -  \\ 
Qwen-Image Edit~\cite{wu2025qwen}    & 0.633 & 0.745  &\underline{0.289} & \underline{133.27} & 0.493 & 0.275 & 155.05 & - & - & - \\ 
USO~\cite{wu2025uso}                 & 0.617 & 0.720  & 0.285  & 147.24 & \underline{0.516} & 0.275 & 141.34 & - & - & -  \\ 
UnicAdapter~\cite{duan2025unic}     & \textbf{0.645} & \textbf{0.762}  & 0.278 & 152.02 & 0.484  & \underline{0.282} & 146.11 & \underline{33.22} & \underline{0.281} & 151.43  \\ 
ControlNet~\cite{zhang2023adding}    & - & - & -  & - & -  & - & - & 28.61 & 0.225 & 187.02  \\ 
UniControl~\cite{qin2023unicontrol}  & - & - & -  & - & -  & - & - & 31.04 & 0.262 & \underline{148.38}  \\ 
\textbf{3SGen}      & 0.638  & \underline{0.757} & \textbf{0.292} & \textbf{132.44} & \textbf{0.520} & \textbf{0.286} & \underline{134.97} & \textbf{35.71} & \textbf{0.298} & \textbf{138.90} \\ 
\bottomrule
\end{tabular}
}
\vspace{-5pt}
\end{table*}

\subsection{Experimental Settings}
\paragraph{Implementation Details.}
We initialize the MLLM from Ming-Omni~\cite{ai2025ming} and implement the connector as a two-layer MLP. The length of the semantic queries is set to $l = 256$ to ensure sufficient semantic capacity, and the size of each memory item is set to $m = 1024$. 
Regarding the training protocol, all three training processes are operated at dynamic resolutions. Both the first and second training stages are each conducted for 3,000 steps, with a learning rate of $1e^{-4}$. In the last stage, the learning rate is further reduced to $3e^{-5}$, and training proceeds for an additional 5,000 steps with a global batch size of 256.

\noindent\textbf{Evaluation Metrics.}
For quantitative evaluation, we assess each task along the following dimensions: (1) subject consistency, quantified by the cosine similarity of CLIP-I~\cite{hessel2021clipscore} and DINO~\cite{oquab2024dinov2} embeddings for subject-driven generation; (2) style similarity, reported via the CSD score~\cite{somepalli2024measuring} for style-driven generation; (3)structure similarity, calculated by the L1 similarity of the structure maps for structure-driven generation; (4) text fidelity, evaluated with CLIP-T~\cite{hessel2021clipscore} across all three tasks; and (5) visual appearance, measured with FID~\cite{heusel2017gans} across all three tasks.

\noindent\textbf{3SGen-Bench.} To enable comprehensive evaluation, we introduce 3SGen-Bench, a unified benchmark comprising content images (including humans, animals and objects), style references (40 style types) and structure maps (Canny, depth, HED, sketch, pose and normal).
Each category contains 100 samples, with representative examples shown in Fig.~\ref{fig:4}.
We further craft prompts for each sample that span a broad range of scenarios, behaviors, and objects. 
For each task, four images per prompt are generated, yielding a total of 1200 samples. 
Given that existing metrics provide insufficient evaluation of image consistency across different categories, we introduce an automated VLM-based metric.
Our quantitative evaluation is based on Qwen2.5-VL~\cite{bai2025qwen2} scoring metrics, assessing both task-specific similarity and adherence to textual instructions.
Additional details are provided in the supplementary material. 
%following GEdit-bench~\cite{liu2025step1x}, 
%The evaluation is based on the score metrics using Qwen2.5-VL~\cite{} following GEdit-Bench~\cite{liu2025step1x}, which assesses the task-specific similarity and instruction consistency of . 

\begin{figure}[t]
    \centering
    %\vspace{-6pt}
    \includegraphics[width=1.0\linewidth]{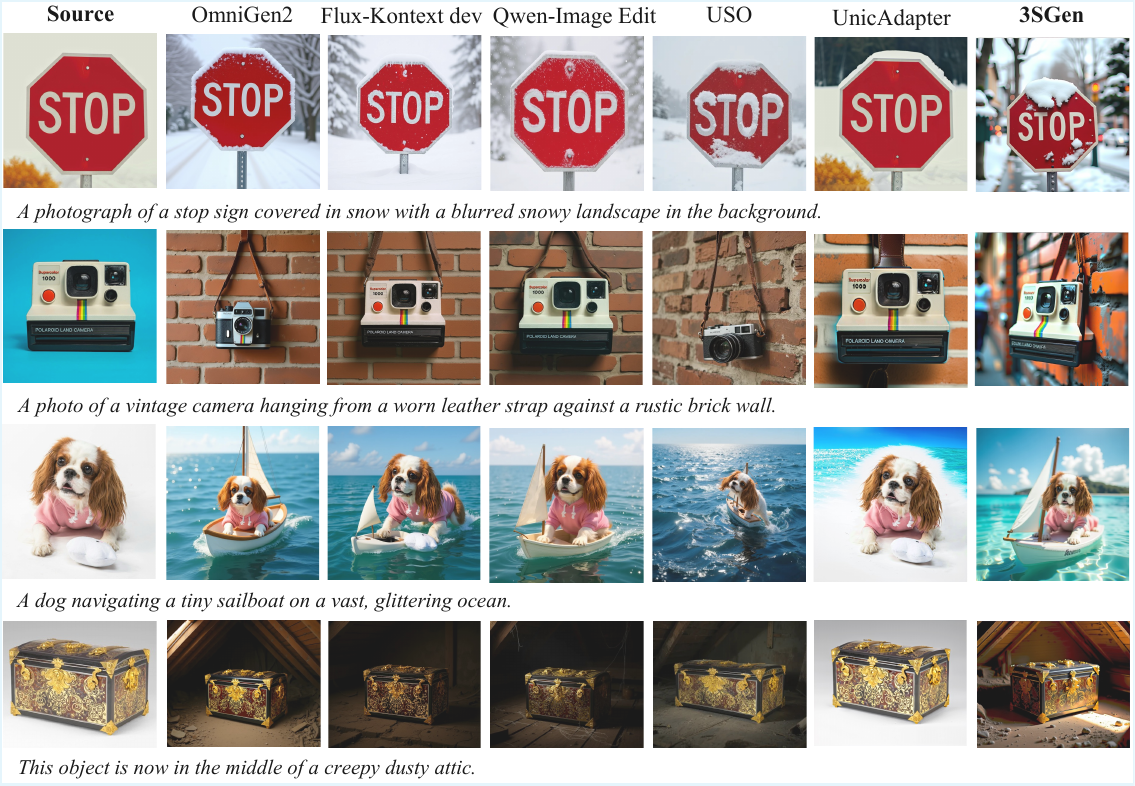}
    \vspace{-15pt}
    \caption{Qualitative comparison of subject-driven generation.}
    \vspace{-10pt}
    \label{fig:5}
\end{figure}

\subsection{Experimental Results}

As a unified customization framework, 3SGen is evaluated against both task-specific and unified baselines. For subject-driven and style-driven generation, we benchmark several unified generation approaches, namely OmniGen2~\cite{wu2025omnigen2}, FLUX Kontext dev~\cite{labs2025flux}, Qwen-Image Edit~\cite{wu2025qwen}, USO~\cite{wu2025uso} and Unic-Adapter~\cite{duan2025unic}. For structure-driven generation, we compare with ControlNet~\cite{zhang2023adding}, UniControl~\cite{qin2023unicontrol} and Unic-Adapter~\cite{duan2025unic}. For the joint style-subject-driven settings with dual conditioning, we compare with USO. Given that existing methods do not support simultaneous structure and style inputs, we adopt the collaboration between UnicAdapter and USO as a baseline, leveraging their respective structure-driven and style-driven capabilities.

\begin{figure}[t]
    \centering
    %\vspace{-6pt}
    \includegraphics[width=1.0\linewidth]{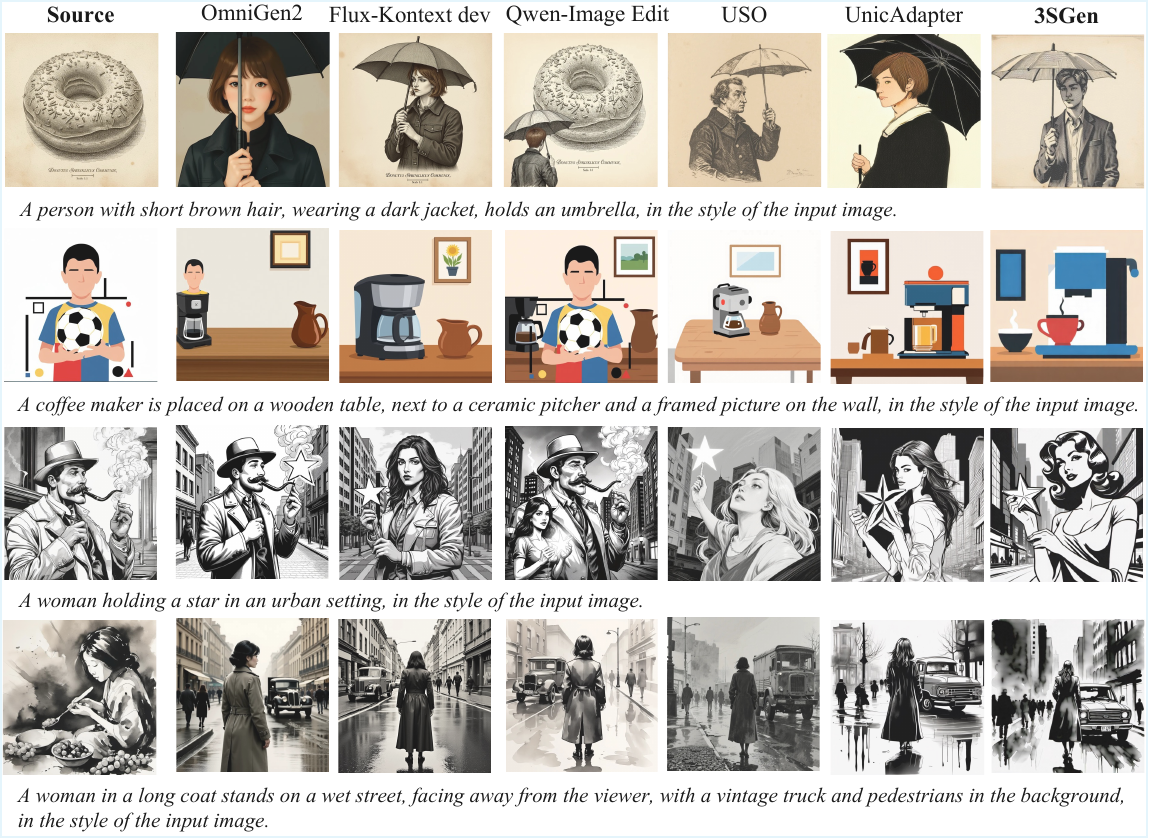}
    \vspace{-15pt}
    \caption{Qualitative comparison of style-driven generation.}
    \vspace{-10pt}
    \label{fig:6}
\end{figure}

\subsubsection{Single-source Driven Generation}
\noindent\textbf{Subject-Driven Generation.}
Tab.~\ref{table:1} presents a quantitative comparison with state-of-the-art methods across all tasks. 
3SGen achieves optimal or near-optimal scores in almost all dimensions, validating its capacity to maintain superior multimodal text-image consistency and visual fidelity across tasks. 
For subject-driven generation, 3SGen ranks slightly behind UnicAdapter and Flux-Kontext in CLIP-I and DINO scores, primarily due to the ``copy-paste" effect evident in the outputs from these baselines, which yields less natural results with limited diversity. 
As illustrated in Fig.~\ref{fig:5}, the results of both UnicAdapter and Flux-Kontext tend toward image-editing paradigms, where differences from the reference images mainly lie in background replacement, leading to degraded text consistency and visual fidelity, as reflected in the metrics reported in Tab.~\ref{table:1}. 
In contrast, 3SGen produces more realistic, diverse, results while preserving strong subject consistency.

\noindent\textbf{Style-Driven Generation.}
Visualization results for style-driven generation are shown in Fig.~\ref{fig:6}. Both OmniGen2 and Qwen-Image Edit retain subject content in their outputs, primarily because these methods simultaneously learn different types of conditional priors without effective feature disentanglement mechanisms, leading to coupling and conflicts between subject and style, as evidenced in Tab.~\ref{table:1}. In contrast, 3SGen achieves optimal performance in style similarity, subject disentanglement, and text consistency.

\begin{figure}[t]
    \centering
    \vspace{-10pt}
    \includegraphics[width=0.85\linewidth]{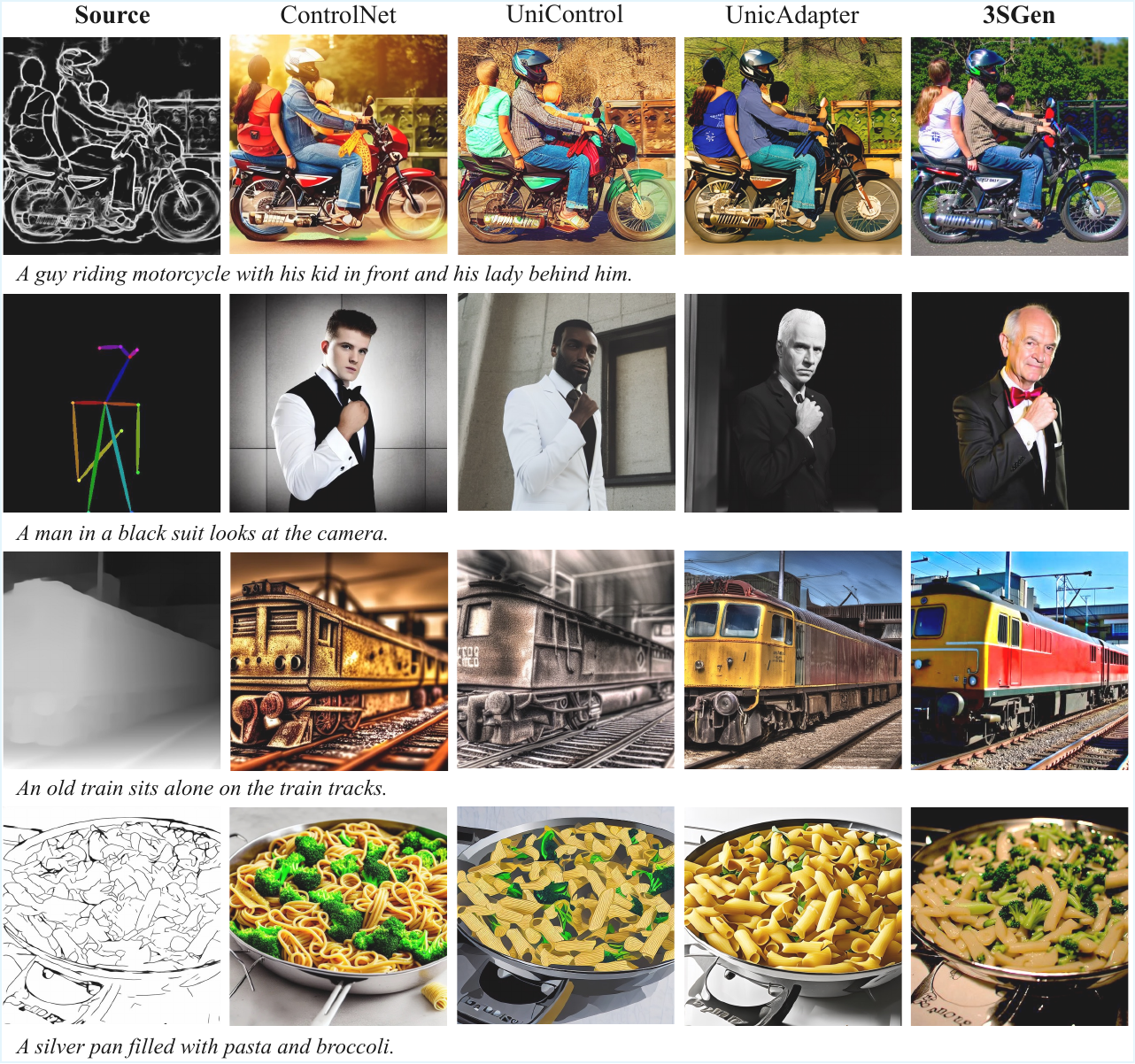}
    \vspace{-4pt}
    \caption{Qualitative comparison of structure-driven generation.}
    \vspace{-5pt}
    \label{fig:7}
\end{figure}

\noindent\textbf{Structure-Driven Generation.}
As shown in Fig.~\ref{fig:7}, 3SGen significantly outperforms existing methods in structure-driven generation. The MLLM, serving as a multi-modal encoder, substantially enriches conditional semantics, while VAE tokens concurrently provide robust spatial structural guidance. Tab.~\ref{table:1} further demonstrates that 3SGen delivers marked improvements in text adherence and visual fidelity.

\begin{table}
    \centering
    %\vspace{-10pt}
    \caption{Quantitative comparison  on our proposed \textbf{3SGen-Bench}.}
    \vspace{-5pt}
    \label{table:2}
    {\footnotesize   %\small
    \begin{tabular}{lcccc}
    \toprule
\multirow{2}{*}{Method} & \multicolumn{4}{c}{Conditional Consistency} \\ \cmidrule(r){2-5} %\cmidrule(r){5}
%& \multirow{2}{*}{Prompt Consistency} 
  & Subject & Style & Structure & Prompt \\ \midrule 
    OmniGen2            & 7.62   & 6.53  &  -  & 8.12  \\
    Flux Kontext dev    & \underline{8.24}   & 7.03   &  - & 8.01  \\
    Qwen-Image Edit     & 7.44   & 6.52   &  - & \underline{8.51}  \\
    USO                 & 6.95   & \underline{7.10}   &  - & 7.35   \\ 
    UnicAdapter         & 6.64   & 5.88   &  7.30 & 6.22  \\ 
    \textbf{3SGen}      & \textbf{8.41} & \textbf{7.35} & \textbf{8.22} & \textbf{8.67}  \\
    \bottomrule
    \end{tabular}
    }
    \vspace{-10pt}
\end{table}

\noindent\textbf{Results on 3SGen-Bench.}
Tab.~\ref{table:2} presents a comprehensive expert evaluation of image-driven generation. 
3SGen consistently outperforms existing methods in both condition alignment on sub-tasks and overall adherence to textual instructions.
Notably, 3SGen achieves an optimal score in subject consistency, which diverges from the CLIP-I and DINO similarity metrics.
This indicates that, rather than performing simply ``copy-paste" operations, 3SGen is capable of generating diverse, coherent, and realistic images.

\begin{figure}[t]
    \centering
    \vspace{-10pt}
    \includegraphics[width=1.0\linewidth]{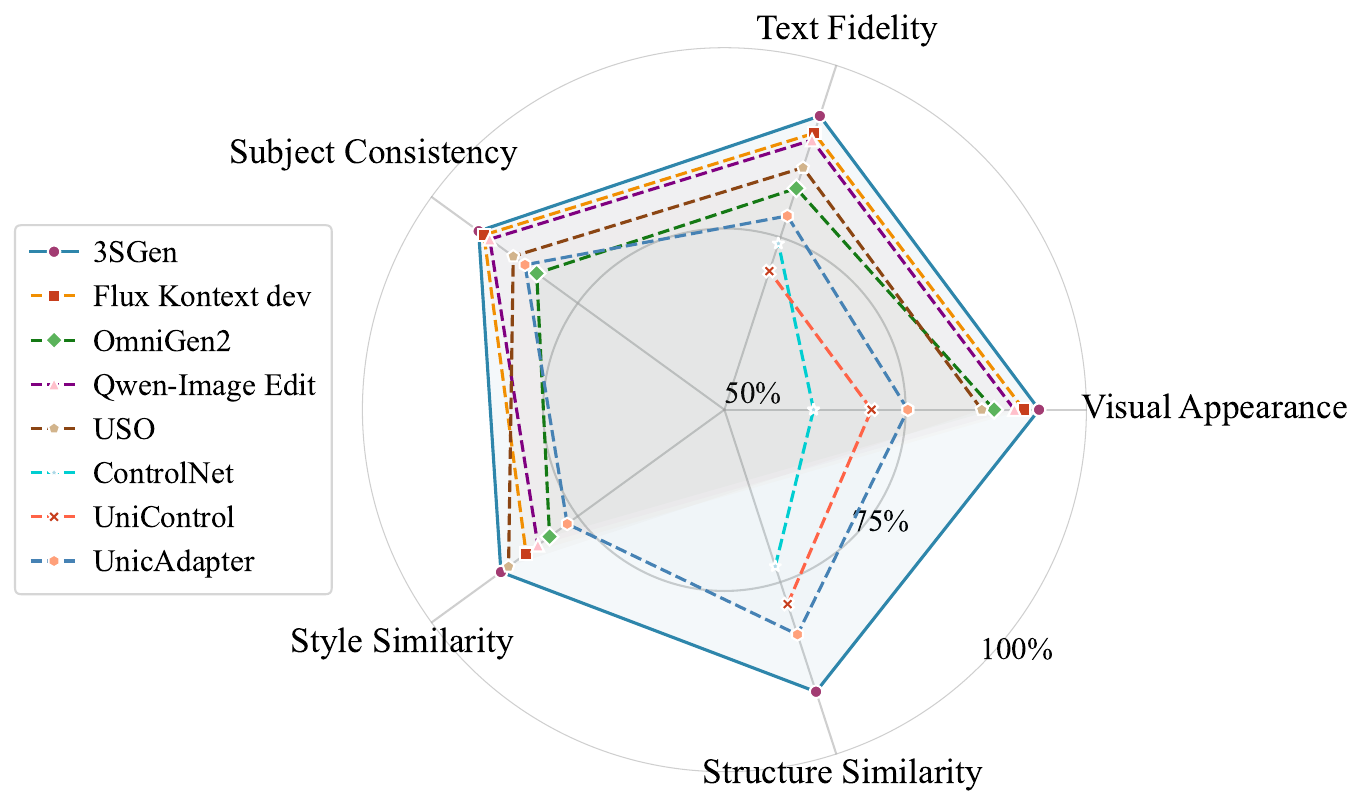}
    \vspace{-12pt}
    \caption{User study of image-driven generation on different dimensions.}
    %\vspace{-12pt}
    \label{fig:8}
\end{figure}

\noindent\textbf{User Study.}
We further conduct user studies to evaluate image-driven generation methods from the perspective of human perception. Participants were asked to rate the results on a scale of 1 to 10 across multiple dimensions, with a total of 85 survey responses collected. As illustrated in Fig.~\ref{fig:8}, 3SGen achieves the highest scores across all dimensions and tasks, demonstrating its superiority and generalizability in cross-task image-driven generation, as well as its excellent visual aesthetics.

\begin{figure}[t]
    \centering
    \vspace{-3pt}
    \includegraphics[width=1.0\linewidth]{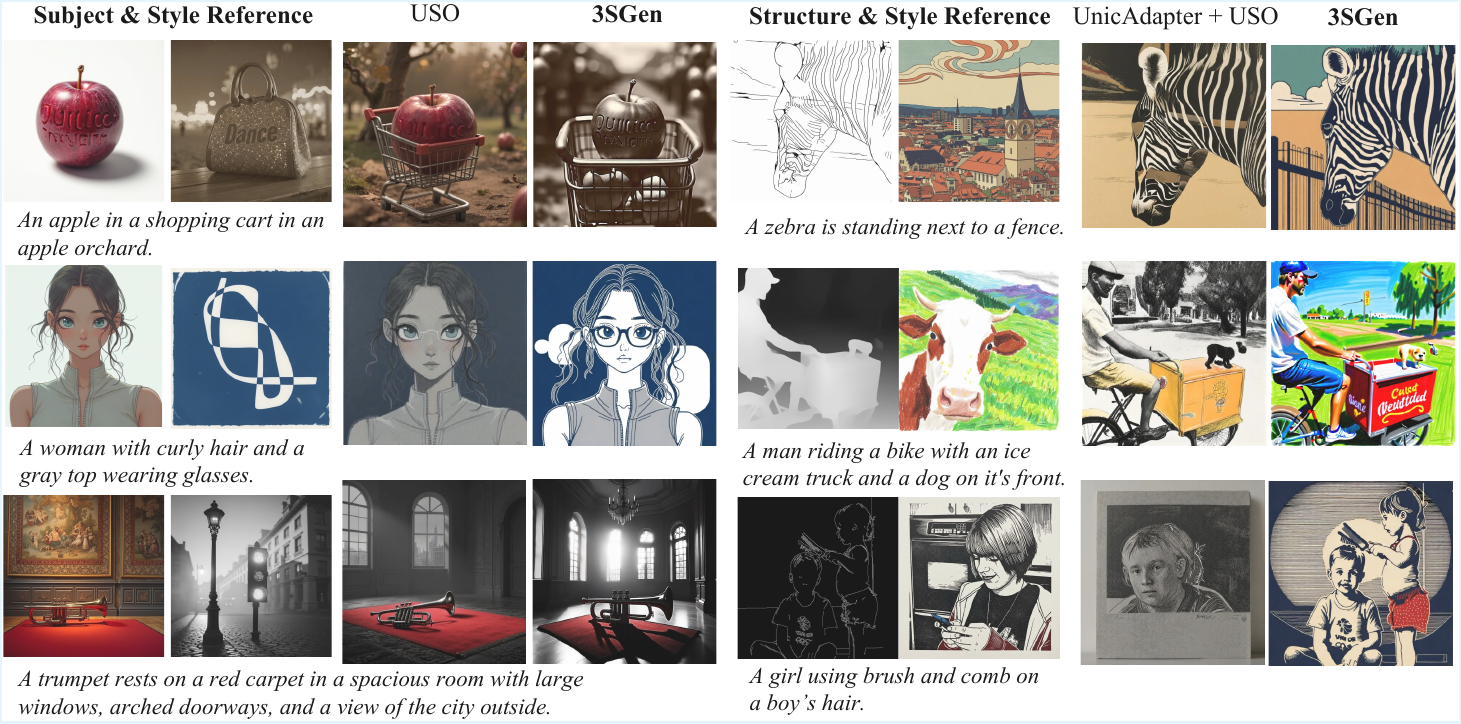}
    \vspace{-12pt}
    \caption{Qualitative comparison of multi-condition compositional generation.}
    \vspace{-12pt}
    \label{fig:9}
\end{figure}

\begin{figure*}
  \centering
  %\vspace{-12pt}
  \includegraphics[width=0.95\linewidth]{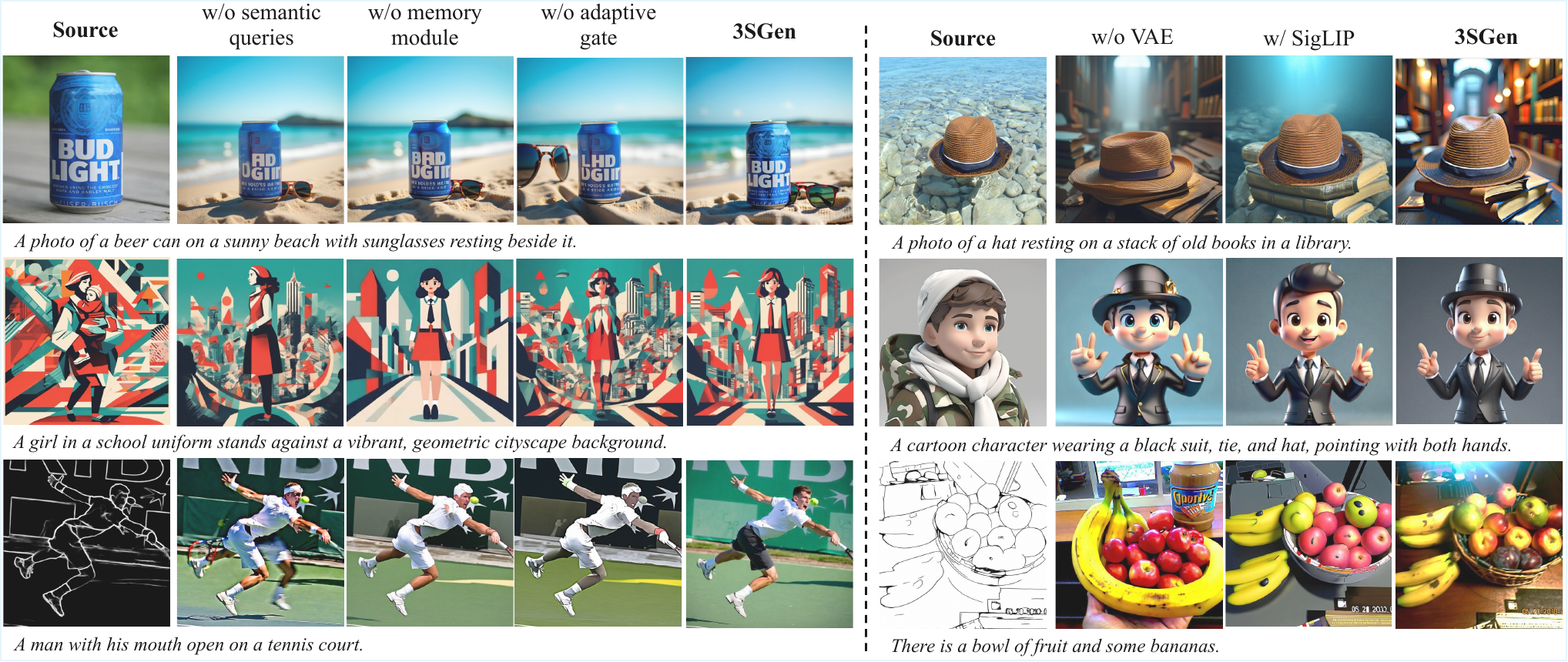}
  \vspace{-5pt}
    \caption{\textbf{Ablation study of our proposed 3SGen.} The left side shows ablations of the ATM module, where ``w/o semantic queries" indicates using the last hidden states of MLLM instead of learnable queries; ``w/o memory module" means directly feeding semantic queries into DiT; ``w/o adaptive gate" represents retrieving arbitrary memory items. The right side presents ablations of visual encoders. ``w/o VAE" denotes the exclusion of VAE embeddings, whereas ``w/ SigLIP" substitutes VAE embeddings with SigLIP embeddings.}
    \vspace{-10pt}
    \label{fig:10}
\end{figure*}

\subsubsection{Multi-source Driven Generation}
We evaluate the performance of multi-condition compositional generation, as shown in Fig.~\ref{fig:9}. For subject and style references, 3SGen exhibits more faithful style reproduction and superior instruction following. In contrast, USO tends to retain the attributes of the original image, failing to achieve balanced consistency across multiple conditions. For structure and style references, 3SGen surpasses the baseline in both structure fidelity and style preservation, demonstrating superior synergistic consistency across multi-source conditions. This further validates our framework's capability for disentangling and unified adaptation of multiple visual attributes.

\subsection{Ablation Study}
\subsubsection{Effect of Adaptive Task-specific Memory Module}

As shown on the left side of Fig.~\ref{fig:10}, replacing semantic queries with the final hidden states of MLLM results in degraded image quality due to semantic deficiency, as evidenced by artifacts in the last row. 
Bypassing the memory module leads to a loss of task-specific conditional priors, which consistently reduces consistency in visual conditions. 
Furthermore, without the adaptive gate, disordered memory item interactions lead to semantic confusion across different conditions, simultaneously affecting visual perception and conditional consistency (text in the first row and style in the second row).
The upper half of Tab.~\ref{table:3} corroborates these observations with quantitative metrics: ``w/o semantic queries" and ``w/o adaptive gate" cause simultaneous declines in both CLIP-T and FID, while ``w/o memory module" primarily induces a decrease in CLIP-I.

\begin{table}
    \centering
    %\vspace{-10pt}
    \caption{\textbf{Ablation study} of the ATM module and vision encoder.}
    \vspace{-5pt}
    \label{table:3}
    {\small
    \begin{tabular}{lccc}
    \toprule
    Method   & CLIP-I $\uparrow$  & CLIP-T $\uparrow$ & FID $\downarrow$ \\
    \midrule
    w/o semantic queries    & 0.615   & 0.265  &  162.13     \\
    w/o memory module  & 0.608 & \underline{0.287}  & 143.09\\
    w/o adaptive gate  &  0.597  & 0.260  & 172.58  \\ 
    \midrule
    w/o VAE Encoder  & \underline{0.618} & 0.263 & 140.06 \\ 
    w/ SigLIP Encoder   & 0.590  & 0.278  &  \underline{139.64} \\ 
    \midrule
    \textbf{3SGen}  & \textbf{0.633} & \textbf{0.295} & \textbf{136.94}   \\
    \bottomrule
    \end{tabular}
    }
    \vspace{-10pt}
\end{table}

To further verify the effectiveness of our task-specific memory, we visualize t-SNE plots for task-specific semantic queries and different memory items. 
Each embedding is projected into a two-dimensional space following global pooling across spatial dimensions.
As illustrated in Fig.~\ref{fig:11}, the feature spaces of different semantic queries exhibit superior organization after modulation, with their projected results demonstrating strong correlation with corresponding memory items.
Furthermore, distinct embedding clusters display clear distributional differences, confirming the effective role of task-specific memory in feature disentanglement and distribution modulation for different conditions.

\subsubsection{Effect of Visual Encoder}

\begin{figure}[t]
    \centering
    %\vspace{-10pt}
    \includegraphics[width=1.0\linewidth]{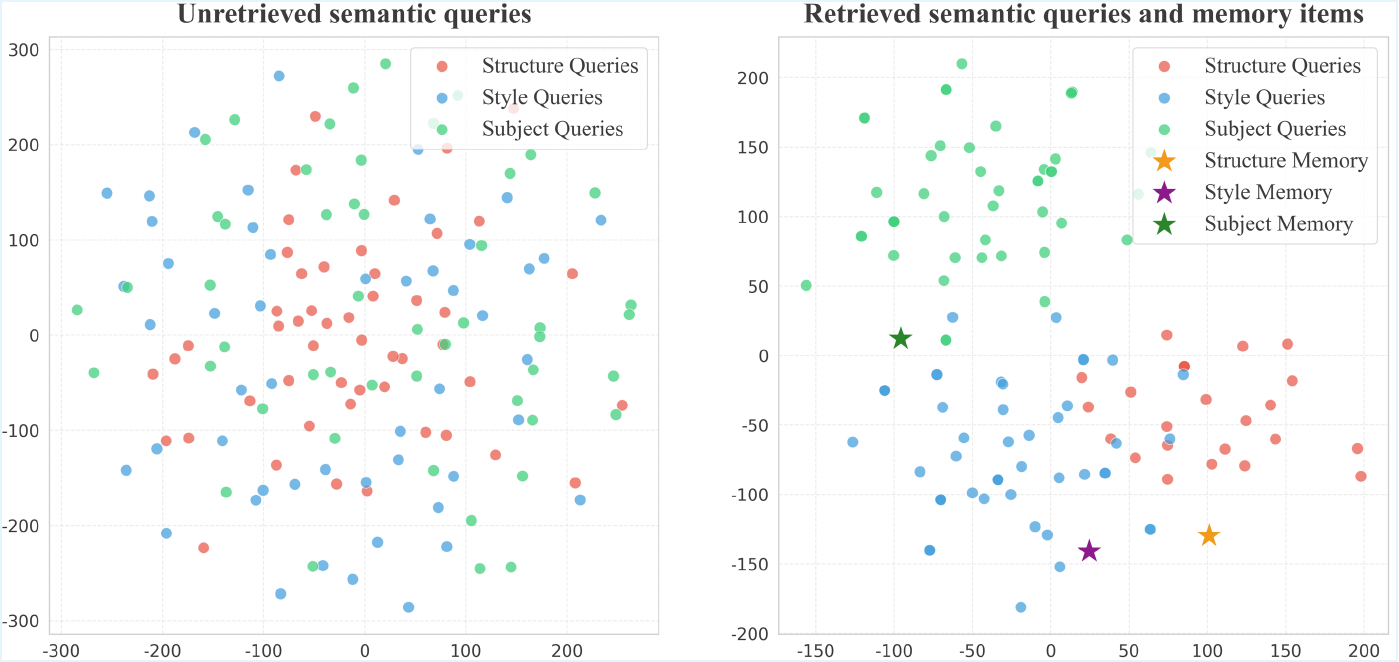}
    \vspace{-12pt}
    \caption{t-SNE visualization of unretrieved semantic queries, retrieved semantic queries and memory items.} 
    %\vspace{-12pt}
    \vspace{-12pt}
    \label{fig:11}
\end{figure}

Removing the VAE encoder leaves only semantic queries as conditional guidance, which not only leads to degraded visual fidelity due to the loss of fine-grained image details, but more critically, impairs structure-driven generation's ability to preserve global spatial layout consistency (the bottom row), as illustrated on the right side of Fig.~\ref{fig:11}. 
On the other hand, replacing the VAE encoder with a SigLIP encoder tends to introduce semantic perturbations, leading to significantly reduced text adherence. For example, the background in the first row and the ``hat" in the second row fail to conform to the prompts. The lower half of Tab.~\ref{table:3} further corroborates these findings quantitatively. In contrast, employing VAE embeddings as visual supplements to semantic queries achieves superior multimodal text-image consistency and excellent aesthetic quality.

\section{Conclusion}
In this work, we introduce 3SGen, a unified framework for image-driven generation that leverages an MLLM with learnable semantic queries to align text–image semantics, complemented by a VAE branch that preserves fine-grained visual details. Central to our approach is an Adaptive Task-specific Memory (ATM) module that dynamically disentangles, stores, and retrieves condition-aware priors across diverse condition types, enabling scalable and consistent multi-task generation within a single model. Extensive experiments on the proposed 3SGen-Bench and public benchmarks demonstrate the superior performance of 3SGen, validating its ability to faithfully synthesize high-fidelity images guided by general multimodal conditions. 
%We hope this framework establishes a solid foundation for future research and provides valuable insights for advancing general-purpose content creation.
{
    \small
    \bibliographystyle{ieeenat_fullname}
    \bibliography{main}
}


% {
%     \small
%     \bibliographystyle{ieeenat_fullname}
%     \bibliography{main}
% }

\end{document}